\newcommand{\modelname}{\emph{DAT}\xspace}
\newcommand{\dcl}{\emph{Depth-aware Negative Suppression loss}\xspace}
\newcommand{\sdcl}{\emph{DNS}\xspace}
\newcommand{\dasca}{\emph{Depth-Aware Spatial Cross-Attation}\xspace}

\def\paperTitle{Introducing Depth into Transformer-based 3D Object Detection}
\def\authorBlock{
    Hao Zhang$^{1,3}$\thanks{This work was done during the internship at IDEA. } \thanks{Equal contribution. }  \qquad
    Hongyang Li$^{2,3}$\footnotemark[1] \footnotemark[2]  \qquad
    Ailing Zeng$^{3}$ \qquad
    Feng Li$^{2,3}$ \\
    Shilong Liu$^{4,3}$\qquad
    Xingyu Liao$^{3}$\qquad
    Lei Zhang$^{1,3}$\thanks{Corresponding author.} \\
    $^1$ The Hong Kong University of Science and Technology.\\
    $^2$ South China University of Technology. \\
    $^3$International Digital Economy Academy (IDEA). \\
    $^4$Dept. of CST., BNRist Center, Institute for AI, Tsinghua University. \\
    $^5$University of Science and Technology of China.\\
}

\newif\ifreview 
\newif\ifarxiv 
\newif\ifcamera 
\newif\ifrebuttal

\documentclass[10pt,twocolumn,letterpaper]{article}
\usepackage{graphicx}
\usepackage{amsmath}
\usepackage{amssymb}
\usepackage{booktabs}

\usepackage{times}
\usepackage{microtype}
\usepackage{epsfig}
\usepackage[table,xcdraw]{xcolor}
\usepackage{caption}
\usepackage{float}
\usepackage{placeins}
\usepackage{color, colortbl}
\usepackage{stfloats}
\usepackage{enumitem}
\usepackage{tabularx}
\usepackage{xstring}
\usepackage{multirow}
\usepackage{xspace}
\usepackage{url}
\usepackage{subcaption}
\usepackage{xcolor}
\usepackage[hang,flushmargin]{footmisc}

\ifcamera \usepackage[accsupp]{axessibility} \fi

% ...

\ifarxiv  \fi

\newcommand{\R}[1]{{%
    \textbf{%
        \ifstrequal{#1}{1}{\textcolor{red}{R#1}}{%
        \ifstrequal{#1}{2}{\textcolor{blue}{R#1}}{%
        \ifstrequal{#1}{3}{\textcolor{magenta}{R#1}}{%
        \ifstrequal{#1}{4}{\textcolor{teal}{R#1}}{%
                           \textcolor{cyan}{R#1}%
        }}}}%
    }%
}}

\usepackage{xr-hyper}
\usepackage{iccv}
\usepackage{times}
\usepackage{epsfig}
\usepackage{graphicx}
\usepackage{amsmath}
\usepackage{amssymb}

\usepackage[pagebackref=true,breaklinks=true,letterpaper=true,colorlinks,bookmarks=false]{hyperref}

\def\iccvPaperID{7157} 

\iccvfinalcopy
\ificcvfinal\fi

\begin{document}

\title{\paperTitle}
\author{\authorBlock}
\maketitle

\begin{abstract}
In this paper, we present \modelname, a Depth-Aware Transformer framework designed for camera-based 3D detection. Our model is based on observing two major issues in existing methods: large depth translation errors and duplicate predictions along depth axes. To mitigate these issues, we propose two key solutions within \modelname. To address the first issue, we introduce a Depth-Aware Spatial Cross-Attention (DA-SCA) module that incorporates depth information into spatial cross-attention when lifting image features to 3D space. To address the second issue, we introduce an auxiliary learning task called \dcl (\sdcl). First, based on their reference points, we organize features as a Bird's-Eye-View (BEV) feature map. Then, we sample positive and negative features along each object ray that connects an object and a camera and train the model to distinguish between them. The proposed DA-SCA and \sdcl methods effectively alleviate these two problems. We show that \modelname is a versatile method that enhances the performance of all three popular models, BEVFormer, DETR3D, and PETR. Our evaluation on BEVFormer demonstrates that \modelname achieves a significant improvement of +2.8 NDS on nuScenes \texttt{val} under the same settings. Moreover, when using pre-trained VoVNet-99 as the backbone, \modelname achieves strong results of 60.0 NDS and 51.5 mAP on nuScenes \texttt{test}. Our code will be released after the blind review.

\end{abstract}
\section{Introduction}
\label{sec:intro}
3D object detection is fundamental for many applications, such as autonomous driving and robotics. Recently, camera-based 3D perception has attracted great attention from both academia and industry. Different from LiDAR-based methods, camera-based methods have advantages such as low cost, long perception range, and the ability to identify vision-only signals such as traffic lights and stop signs. One key challenge for camera-based methods is the lack of depth information. Previous works \cite{li2022bevdepth,dd3d} have proved that high-quality depth can remarkably help improve 3D detection performance. Although many works try to recover depth from camera images, depth estimation is still an ill-posed problem. 

\begin{figure}[tp]
    \centering
    \includegraphics[width=\linewidth]{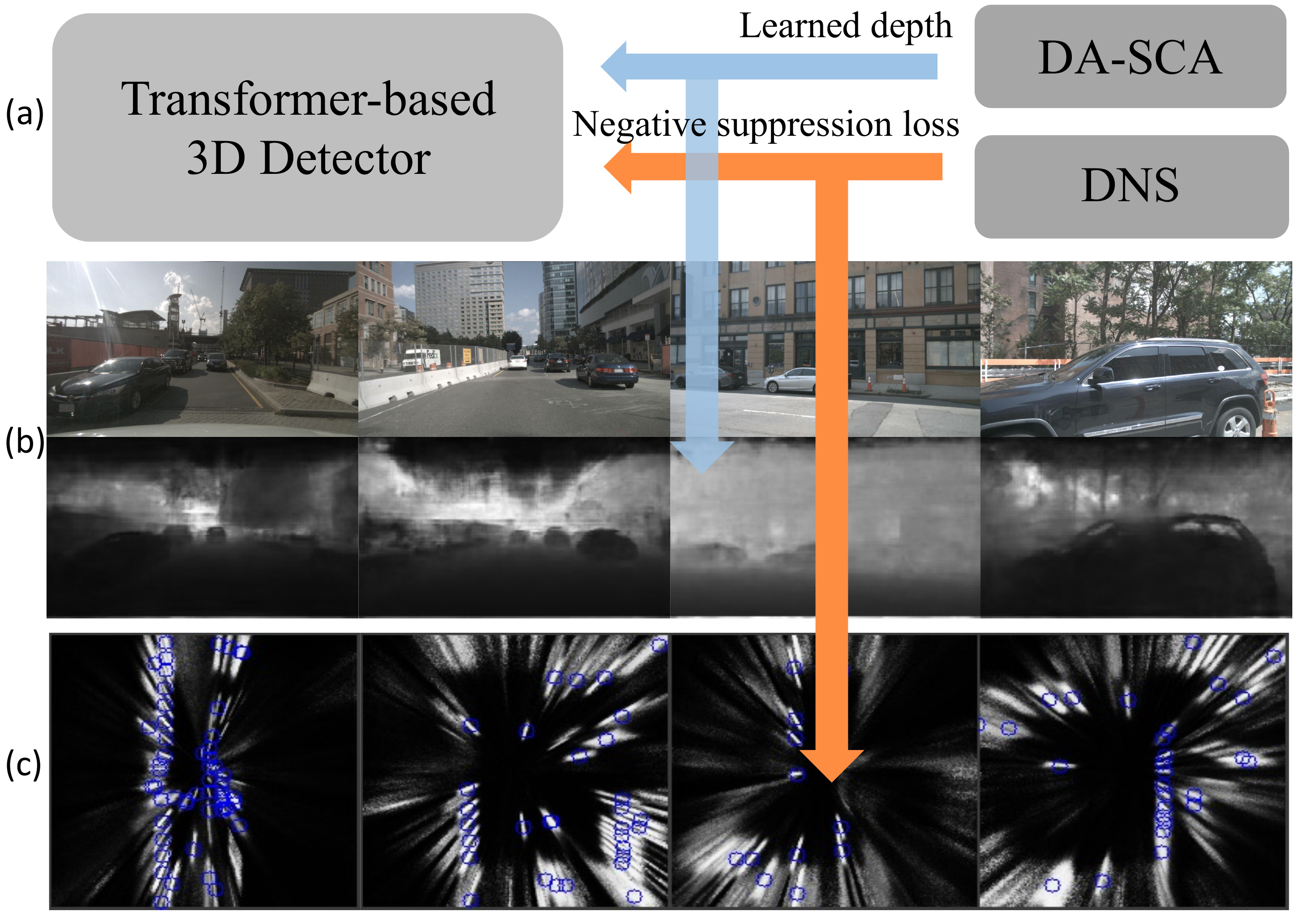}
    \caption{ (a) Our method introduces both explicit and implicit depth to 3D Transformer. (b) A visualization of our depth map learned in Depth-Aware Spatial Cross-Attention (DA-SCA). The upper row shows the input images and the lower row gives the corresponding depth maps where higher brightness denotes a deeper distance. (c) A visualization of "objectiveness scores" predicted by our model where brightness denotes the confidence score that there is an object. The blue circles are the centers of ground truth (GT) objects.
    }
    \label{fig:intro}
    \vspace{-5mm}
\end{figure}

Some works explicitly estimate the depth and then use the predicted depth for downstream tasks. For example, a branch of research ~\cite{xu2018multi,weng2019monocular,wang2019pseudo} utilizes a stand-alone depth estimation network to output depth maps and generates pseudo-LiDAR information from the depth maps. 
Some researchers~\cite{reading2021categorical,chen2020dsgn} also find this paradigm sub-optimal due to non-differentiable transformation from 2D to 3D space. Recently, BEVDepth~\cite{li2022bevdepth} proposes to supervise depth learning with LiDAR-generated sparse depth labels. It also leverages camera information for more precise depth estimation and proposes a depth correction module to reduce error from disturbance of camera extrinsic. BEVDepth obtains a strong performance due to the high quality of depth it estimates.  Another line of work focuses on directly predicting 3D boxes without predicting depth. For example, researchers in~\cite{zhou2019objects} found that CenterNet can predict 3D boxes by slightly modifying the box head. 
This line of work can be viewed as implicitly learning depth by predicting the 3D locations of object boxes. More recently, some Transformer-based 3D detectors have been proposed~\cite{wang2022detr3d, liu2022petr, li2022bevformer} motivated by Transformer-based models for 2D object detection. Although they are among the top-performing camera-based 3D detectors, they all share a common problem when using object queries to probe image features.

The problem arises from conducting Spatial Cross-Attention (SCA)\footnote{We follow BEVFormer and refer to the cross-attention that probes image features to Bird's-Eye-View (BEV) features as Spatial Cross-Attention.} in a way that is naively inherited from 2D detection without considering the intrinsic characteristics of 3D space. Specifically, they only take $(u, v)$\footnote{By default, we use $(u, v, d)$ to denote coordinates in a camera coordinate system, where $(u, v)$ are the 2D coordinates in an image plane and $d$ is the depth whose direction is perpendicular to the image plane. We also use $(x, y, z)$ to denote coordinates in an ego vehicle coordinate system.} into consideration and ignore the crucial information provided by $d$. Without depth, their approach is conceptually indirect and inefficient for the detection head to distinguish the depths of different objects. As a result, two problems occur (see Section~\ref{sec:motivation}): predictions have relatively large errors along depth axes, and ambiguity leads to redundant predictions along depth axes.
To address the two problems, this paper proposes two novel improvements, as shown in Fig.\ref{fig:intro}. The first improvement involves introducing depth into the Spatial Cross-Attention (SCA) module. For each object query, when probing features from a camera view, the positional part is the positional encoding of $(u, v, d)$ converted from its 3D reference point $(x, y, z)$ based on the BEV-to-camera transformation matrix. Additionally, $(u, v, d)$ encoding is added to image features, where $d$ is predicted from image features with a depth network trained with LiDAR-generated depth labels similar to BEVDepth\cite{li2022bevdepth}. Fig.~\ref{fig:intro}(b) shows depth maps predicted by the proposed method, which indicate that depth can be obtained with good quality with the depth network. The results in Table~\ref{tab:abl:components} show that \dasca effectively improves detection performance.

For the redundant prediction problem, we propose a \dcl (\sdcl) approach to help the network learn to reject predictions with incorrect depths and reduce redundant predictions. Specifically, positive and negative BEV features are sampled along a ray from a camera to an object center, denoted as an object ray, as shown in Fig.\ref{fig:motivation bevformer}(b). The BEV features at object centers are considered positive, while other BEV features on object rays are negative. An auxiliary loss is applied to the sampled BEV features in the training phase, where positive features are expected to have high-class scores and negative ones are expected to have low-class scores. This approach effectively encourages the network to suppress redundant predictions and only keep the most accurate prediction on each object ray. The visualization of class scores in Fig.\ref{fig:intro} (c) shows that our model can distinguish the correct depth of objects. Note that the brightness in Fig.\ref{fig:intro} (c) denotes the class score, and blue circles represent the BEV positions of GT boxes. Although the \sdcl is applied to BEV queries, the proposed method is generic and can be applied to non-BEV methods such as DETR3D and PETR. For these methods, the output features of spatial cross-attention should first be organized as a sparse BEV feature map, as shown in Fig.\ref{fig:arch}(b). When sampling a BEV feature at coordinates $(x, y)$, a feature can be directly sampled if there is any feature near $(x, y)$ (within a predefined scope). Otherwise, a pseudo-BEV query with a reference point corresponding to $(x, y)$ can be constructed.
In summary, this paper proposes three key contributions. Firstly, a Depth-Aware Spatial Cross-Attention (DA-SCA) module is proposed, which mitigates an important issue of high translation errors along depth in Transformer-based camera-only 3D detectors by incorporating depth positional encoding into both query and image features during spatial cross-attention. Secondly, the paper introduces a \dcl (\sdcl) to minimize redundant predictions on object rays, thereby significantly improving performance. Finally, comprehensive experiments on the nuScenes dataset validate the proposed method's effectiveness, demonstrating significant performance gains on several Transformer-based 3D detectors such as BEVFormer, DETR3D, and PETR. When applied to BEVFormer, the proposed approach achieves a 54.5 NDS score with ResNet101 as the backbone, which is a +2.8 NDS improvement over BEVFormer. Additionally, the paper achieves a strong result of 60.0 NDS with VoVNet-99 pre-trained on a dense depth estimation task as the backbone.
\section{Related Work}
\label{sec:related}
\noindent\textbf{Depth learning in camera-based 3D detection.} 
Camera-based 3D detection requires predicting 3D boxes only from camera data, in which depth learning is the most challenging because it is an ill-posed problem. One line of works predicts depth explicitly and then uses depth for 3D detection. These works can be divided into two categories~\cite{qian20223d}: result-lifting methods and feature-lifting methods. Result-lifting methods\cite{chabot2017deep,mousavian20173d,li2019gs3d,li2019stereo,shi2021geometry} break down 3D detection into 2D detection and depth prediction and predict objects according to geometric properties and constraints. These methods severely rely on feature engineering which hinders them from generalizing to various scenarios. Feature lifting methods lift image features into 3D space. Some~\cite{you2019pseudo,park2021pseudo,wang2019pseudo} predict depth map from image features and then lift the depth map into pseudo-LiDAR to mimic LiDAR signal. Once the pseudo-LiDAR signal is generated, they can utilize a LiDAR-based detection head for detection. Some learn a categorical depth distribution~\cite{li2022bevdepth,reading2021categorical} and then project image features into 3D space~\cite{reading2021categorical} or BEV plane~\cite{li2022bevdepth}. BEVDepth~\cite{li2022bevdepth} supervises depth learning with sparse LiDAR data and proposes a depth correction module to reduce depth error. Their method establishes a new SOTA result on nuScenes 3D detection task with camera data only. The explicit depth learning methods require a depth network and ground truth depth labels. Another branch of research learn depth implicitly. For example, DETR3D uses object queries to probe features and output a prediction for each query without predicting depth.
\\
\noindent\textbf{Transformer-based camera-only 3D detectors.}
Recently, Transformer-based models~\cite{carion2020end,zhu2020deformable,liu2021dab,li2022dn,zhang2022dino} have achieved great success in 2D detection. The core of the Transformer-based models is to query image features with some object queries, which is called Spatial Cross-Attention (SCA) in this paper. The first Transformer-based model on nuScenes detection task is DETR3D~\cite{wang2022detr3d}. It contains a feature extractor and a detection head. The detection head converts 3D reference points into camera reference points and then uses single-point deformable attention to probe image features. PETR~\cite{liu2022petr} is also a simple DETR framework for 3D detection. It adopts standard attention that uses the relativity of key and query to guide feature probing. Instead of mapping 3D reference points to 2D reference points in the camera plane, it lifts the camera plane into 3D space along depth axes. Because an image feature only has camera coordinates $(u, v)$ and does not contain depth. It samples depths with equal space in a valid scope denoted as ${d_1, d_2, ...,d_k}$. For each depth $d_i$, it converts $(u, v, d_i)$ into 3D coordinates $(x_i, y_i, z_i)$. The positional encoding of the key is the sum over positional encodings of all $(x_i, y_i, z_i)$ for i from 1 to k. Therefore, the key and query are aligned in an ego-vehicle coordinate system. Another work, BEVFormer~\cite{li2022bevformer}, is similar to DETR3D, converting 3D reference points into camera reference points. It uses BEV queries to map image features into the BEV plane and utilizes an extra detection head to predict boxes. In summary, DETR3D and BEVFormer map 3D reference point into the camera plane without using depth information. PETR lifts image features along depth axes, but it is not biased on depths.

\section{Method}
\label{sec:method}
\subsection{Motivation}
\label{sec:motivation}
\begin{figure}[tp]
    \centering
    \includegraphics[width=\linewidth]{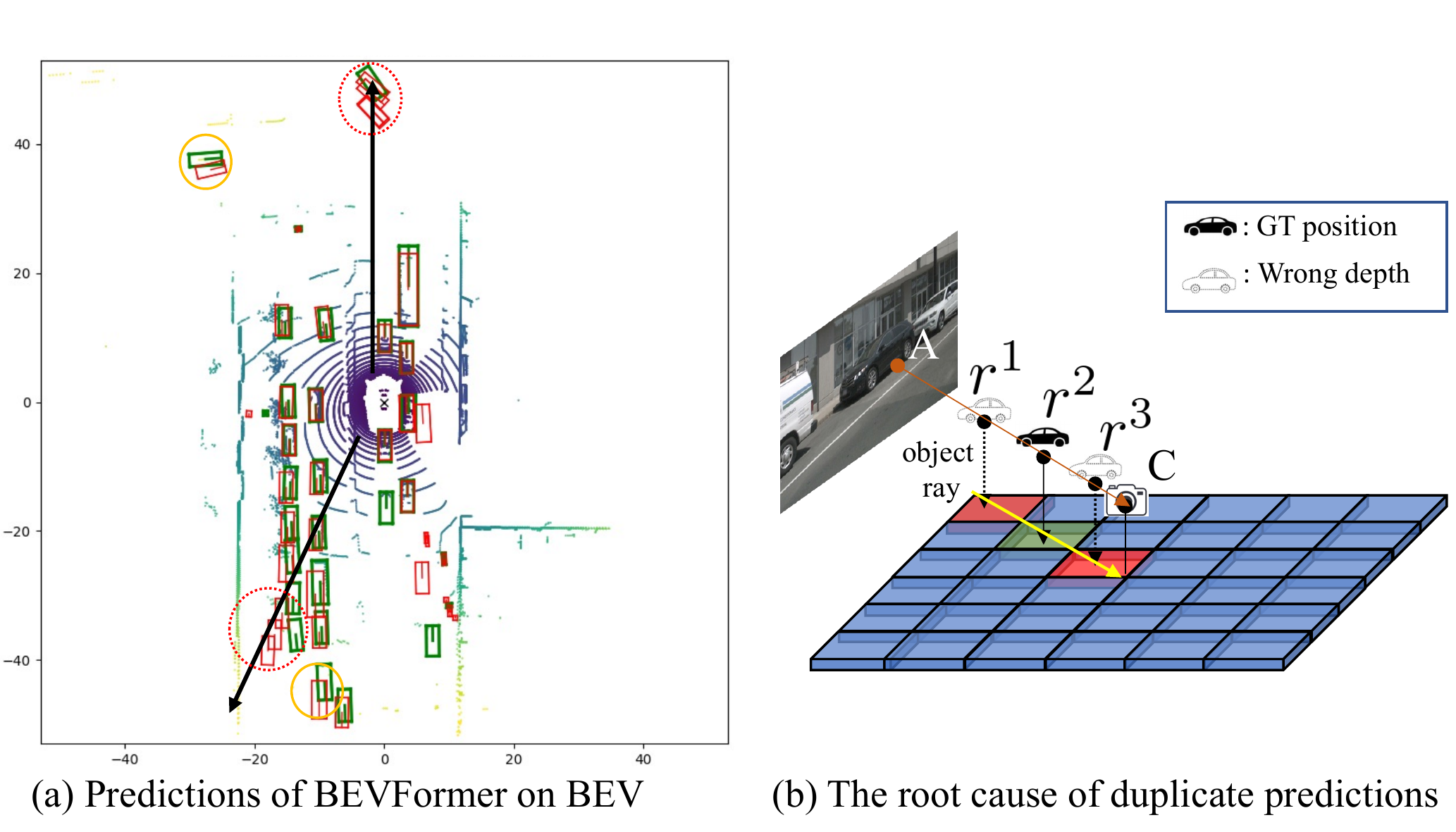}
    \caption{(a) Visualizations of the predicted boxes of BEVFormer on Bird's Eye View (BEV). Green and red boxes denote GT and predicted boxes, respectively. (b) illustrates the redundant prediction problem along the object rays represented with the yellow arrow line.}
    \label{fig:motivation bevformer}
    \vspace{-0.7cm}
\end{figure}

%
Our approach was motivated by observing two failure modes in Transformer-based 3D detectors, such as BEVFormer. The first failure mode is that these models often have large translation errors along depth axes, as indicated by the yellow circles in Fig.\ref{fig:motivation bevformer}. To address this issue, one intuitive solution is to introduce depth as BEVDepth does. BEVDepth predicts a depth map from the image and then employs an Lift Splat Shoot (LSS)-based~\cite{philion2020lift} feature lifting approach and a voxel pooling module to obtain a BEV embedding. However, our method is a Transformer-based detector that uses cross-attention to query image features into 3D space. Therefore, we propose a Depth-Aware Spatial Cross-Attention module to incorporate depth into cross-attention, as detailed in Section\ref{subsec:sca}.

The second failure mode is that these detectors often predict duplicate objects along the object ray, as indicated by the red dashed circles in Fig.\ref{fig:motivation bevformer}. We use BEVFormer as an example to illustrate the reason for this problem in Fig.\ref{fig:motivation bevformer}(b). In the Spatial Cross-Attention module of BEVFormer, a 3D reference point $r^i=(x^i, y^i, z^i)$ is mapped to a camera reference point $(u^i, v^i)$ on a camera plane, which is then treated as a 2D reference point for deformable attention. However, when there are multiple 3D reference points $r^1$, $r^2$, and $r^3$ along an object ray with corresponding BEV features $q^1$, $q^2$, and $q^3$, the queries in different 3D locations such as $q^1$, $q^2$, and $q^3$ will probe image features based on the same camera reference point $(u^1, v^1)=(u^2, v^2)=(u^3, v^3)$. This ambiguity makes it challenging for the detection head to distinguish the object's exact location, particularly along the depth direction, and leads to duplicate predictions. We argue that this is a common problem in current Transformer-based 3D detectors, including DETR3D~\cite{wang2022detr3d} and PETR~\cite{liu2022petr}, which query image features similarly. To solve this problem, we propose a \dcl approach, referred to as \sdcl, which focuses on distinguishing hard negative positions from the ground truth positions. The details of our approach are described in Section~\ref{subsec:DCL}.
\subsection{Overview}
\label{subsec:Overview}
\begin{figure*}[tp]
    \centering
    \includegraphics[width=0.9\linewidth]{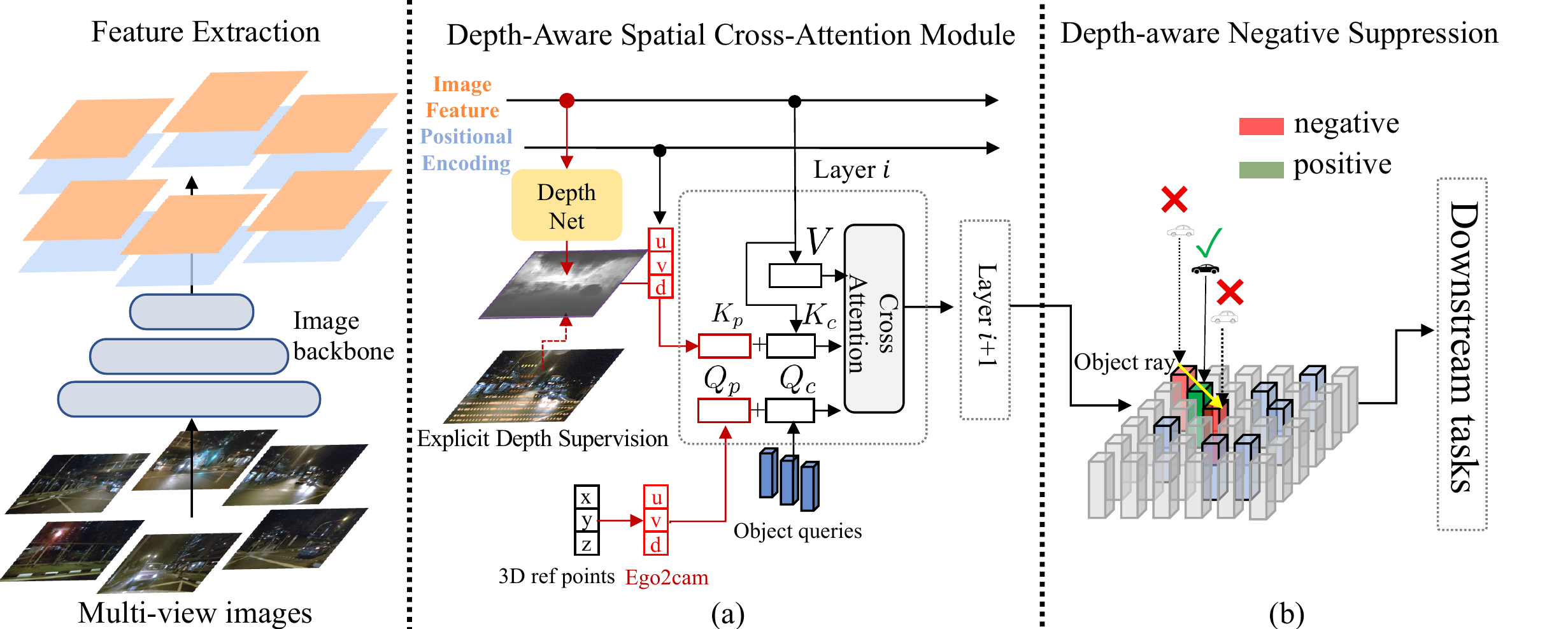}
    \caption{The training pipeline of our method consists of four parts: the feature extractor, \dasca (DA-SCA), \dcl(\sdcl), and the downstream task. (a) shows the DA-SCA, which includes a depth network and a cross-attention module. Our proposed components are indicated with red lines. The depth network takes the image features as input and outputs a depth map, which is encoded as Positional Encodings (PE) together with $(u, v)$ PE of the images. This depth-aware PE serves as the positional key of spatial cross-attention. On the query side, there are object queries and 3D reference points. The 3D reference points are either predicted by the object queries, as in DETR3D and PETR, or pre-defined coordinates, as in BEVFormer. The $(u, v, d)$ coordinates can be obtained through ego-to-camera transformation matrices and are encoded as positional queries. In (b), we place the output features of the DA-SCA on a BEV plane according to the $(x, y)$ coordinates of their 3D reference points. The colored features in the figure represent actual features, while the uncolored ones are placeholders included for illustrative purposes. Note that for BEVFormer, the BEV feature is dense and does not require placeholders. The yellow arrow connecting the center of an object (green feature) and a camera on the BEV features is referred to as an object ray. The object ray intersects with multiple BEV queries, where the positive one is shown in green and the negative ones are shown in red.}
    \label{fig:arch}
    \vspace{-2mm}
\end{figure*}

Our training pipeline consists of three parts, as illustrated in Fig.~\ref{fig:arch}. First, the feature extractor takes multi-view images as input and produces image features and positional encodings used by the DA-SCA module. Second, DA-SCA utilizes image features to predict a depth map using a depth network. Like BEVDepth, the depth network is trained with LiDAR-generated depth labels during training. It is worth noting that we do not use any ground truth depth information during inference. The learned depth is encoded as a positional key and the positional encoding of $(u, v)$ of image features. On the query side, 3D reference points are converted to $(u, v, d)$ coordinates on the image plane by multiplying the ego-to-camera transformation matrix. The positional encoding of $(u, v, d)$ coordinates is then encoded into queries. Finally, the output features from DA-SCA can be used for downstream tasks. Before that, we apply an auxiliary loss, called \dcl, on the output features to improve the model's ability to reject hard negative examples that may lead to duplicate predictions.
\subsection{Depth-Aware Spatial Cross-Attention}
\label{subsec:sca}
\begin{figure}[tp]
    \centering
    \vspace{-2mm}
    \includegraphics[width=\linewidth]{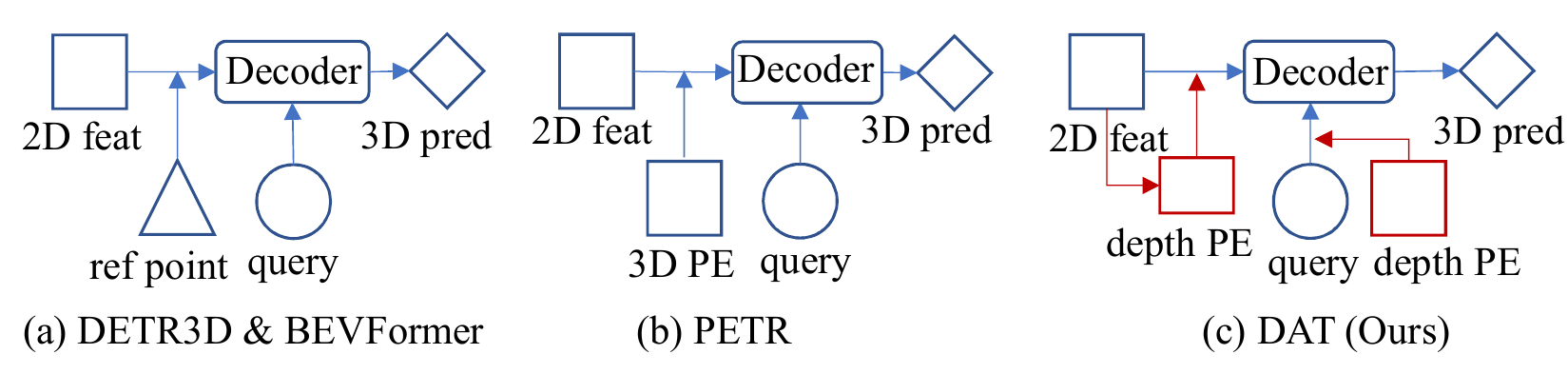}
    \caption{ 
    This figure compares the spatial cross-attention mechanisms of DETR3D/BEVFormer, PETR, and our proposed \modelname. DETR3D and BEVFormer project 3D points to the 2D image plane, while PETR maps 3D positions onto the image plane using 3D position embeddings. Our method, \modelname, includes depth positional encodings in both the query and image features to capture depth information. This allows for more accurate object position prediction and improved detection performance.}
    \label{fig:dasca}
    \vspace{-2mm}
\end{figure}

Previous Transformer-based 3D detectors have ignored depth information when performing spatial cross-attention. In contrast, we have carefully considered depth in this work, as shown in Figure~\ref{fig:dasca}. DETR3D and BEVFormer map reference points onto the image plane and only consider 2D space during spatial cross-attention. PETR proposes to add 3D positional encoding (PE) to image features, but the 3D PE is represented with $(x, y, z)$ coordinates and is obtained by taking averages on all possible depths, providing limited depth information. Our DA-SCA module differs from previous approaches in that the depth in image features is meaningful depth predicted by a depth network. Consequently, we convert 3D reference points into $(u,v,d)$ coordinates and use their PE as the positional query for spatial cross-attention.
To clarify the workings of our DA-SCA module, we denote the image feature map as $\mathbf{F}$ with a shape of ${\begin{matrix} h \times w \times c \end{matrix}}$, the predicted depth as $\mathbf{d}$ with a shape of ${\begin{matrix} h \times w \times 1 \end{matrix}}$, and the 3D reference point as $(x, y, z)$. The positional query $Q_p$ is calculated as follows:

\begin{equation}
\begin{aligned}
u,v,d&=\text{Ego2cam}(x,y,z)\\
Q_p&=\text{SinePE}(u,v,d),
\end{aligned}
\end{equation}

Here, Ego2cam represents the transformation from the ego-vehicle coordinate system to a camera coordinate system, and SinePE is the sine positional encoding function. For a pixel with coordinates $(u_c, v_c)$, the positional key $K_p$ can be obtained as follows:

\begin{equation}
K_p=\text{SinePE}(u_c,v_c,\mathbf{d}[u_c,v_c]),
\end{equation}

where $\mathbf{d}[u_c,v_c]$ is the value of $\mathbf{d}$ at $(u_c,v_c)$. We also have $V=K_c=\mathbf{F}[u_c,v_c]$. Thus, the relation between the input and output of DA-SCA can be expressed as follows:

\begin{equation}
\begin{aligned}
F&=\text{DA-SCA}(Q,K,V)\\
&=\text{DA-SCA}(Q_c+Q_p,K_c+K_p,V)\\
&=\text{DA-SCA}(Q_c+\text{SinePE}(u,v,d),\mathbf{F}[u_c,v_c]+\\ &\quad\text{SinePE}(u_c,v_c,\mathbf{d}[u_c,v_c]),\mathbf{F}[u_c,v_c])
\end{aligned}
\end{equation}

In summary, these equations define how the positional query and key are computed using the predicted depth and the positional encoding functions. Additionally, they show how the input and output features of DA-SCA are related. Our \dasca is effective according to Table~\ref{tab:abl:components}, which improves NDS by $1.0\%$ and reduces mean Average Translation Error (mATE) by $1.3\%$. We show more analysis of how our method works in the Appendix.


\subsection{Depth-Aware Negative Suppression Loss}
\label{subsec:DCL}
To explicitly reduce duplicate predictions, we propose an auxiliary learning task, called \dcl (\sdcl). This task encourages the model to learn to reject hard negatives along object rays, thereby reducing duplicate predictions. We assign an object ray for each object, and each point on the object ray can be mapped to a BEV position. For BEVFormer, which has a dense BEV feature map, we can sample a BEV feature at any position on the BEV. For DETR3D and PETR, which do not have a dense BEV feature map, we construct a pseudo-BEV query with a learnable query and a 3D reference point $(x, y, z)$, where $(x, y)$ is the desired BEV position, and use the pseudo-BEV query to query the feature from the image feature map. We sample the BEV feature at the ground truth position as the positive example and $N_{neg}$ negative examples along the object ray with incorrect depths, denoted as $\left\{{f_1, f_2, ..., f_{N_{pos}+N_{neg}}}\right\}$, since we can sample BEV features at any position.

After the positive and negative features are sampled, they are fed into a classification head $H_{cls}(\cdot)$, and the predictions are utilized for the auxiliary task during training. We supervise the positive BEV features with the ground truth class of the object and suppress class scores of all categories for negative BEV features. The \dcl is formulated as follows:

\begin{equation}
\begin{aligned}
L_{\sdcl}&=\sum_{c \in C}L_{bce}\left(H_{cls}(f_p),y_c\right)\\&+\sum_{i \in I_{neg}}\sum_{c \in C}L_{bce}\left(H_{cls}(f_i),0\right),
\end{aligned}
\end{equation}

where $f_p$ is the positive feature and $I_{neg}$ are the sets of indices of negative BEV features. $C$ is the set of all categories. This loss enforces the model to output high-class scores for positive BEV features on their GT label and low-class scores for negative ones on all categories, which helps the model learn to reject hard negatives and reduce duplicate predictions. Visualization in Fig.~\ref{fig:vis} verifies that our method indeed reduces duplicate predictions.

\section{Experiments}
\label{sec:experiments}
\subsection{Experimental settings}

To ensure that our results are reproducible, we provide details of our experiment settings in datasets metrics, implementation details and how we implement our rescale-based query selection.

\noindent\textbf{Datasets.} Our experiments are conducted on the nuScenes dataset~\cite{caesar2020nuscenes}, which is a large-scale public dataset that contains 1000 scenes, each of 20 seconds in duration. The dataset provides data from various sensors, including cameras, RADAR, LiDAR, and more. Since our task is camera-based 3D detection, we only use camera images as the data source. The entire dataset is annotated at 2Hz, with 40k keyframes being annotated. It is important to note that nuScenes redefines mAP.

\noindent\textbf{Metrics.} In contrast to the mAP used in 2D detection, which employs Intersection over Union (IoU) to match predicted results and ground truth, nuScenes uses center distance for matching. NuScenes also defines five true positive (TP) metrics: ATE, ASE, AOE, AVE, and AAE, for measuring translation, scale, orientation, velocity, and attribute errors, respectively. The main metric used in nuScenes is the nuScenes detection score (NDS), which considers all the TP metrics. We report our results on mAP, NDS, and all the TP metrics.

\noindent\textbf{Implementation details.} \label{implementation}To ensure a fair comparison, we use exactly the same settings as the baseline method we compare with, except for the methods we propose. Note that our baseline method is BEVFormer, except for Table~\ref{tab:abl:univ}. We provide the implementation details of tables other than Table~\ref{tab:abl:univ} in this section, and more details will be provided in the Appendix. We adopt two backbones: ResNet101-DCN\cite{resnet,dai2017deformable} initialized from the FCOS3D\cite{wang2021fcos3d} checkpoint, and VoVNet-99~\cite{lee2019energy} initialized from the DD3D~\cite{dd3d} checkpoint. By default, we use feature maps with sizes $1/16, 1/32$, and $1/32$ with hidden dimensions $256$. The BEV resolution is $200\times200$. In Spatial Cross-Attention, each query corresponds to $4$ reference points with different heights. We also adopt 6 BEV encoder layers, where each contains a Temporal Self-Attention and a Spatial Cross-Attention. Same as BEVFormer, we also adopt a deformable decoder with $6$ layers and $900$ queries as our box head. As for DA-SCA, we use a simple MLP with $2$ Linear layers to predict the depth map. We choose Sine Positional Encoding for both query and value. For \sdcl, we choose 3 positive and 3 negative examples and set the multiplier $\lambda$ mentioned in section~\ref{subsec:DCL} to $0.2$. We train our model with a learning rate $2\times 10^{-4}$ for $24$ epochs by default. \\
\noindent\textbf{Rescale-based Query Selection:} We implemented a two-stage (query selection) DETR head as our detection head. We would like to clarify that this method is not considered a contribution to our work, and we describe it solely for implementation convenience. We find that directly using the query selection proposed by deformable DETR does not work because the resolution of the BEV feature map is too high. There are $200\times 200=40000$ BEV features in total and directly selecting $900$ queries cannot cover all objects. Therefore, we first interpolate the BEV features from $200\times 200$ to $50\times 50$ and then select $900$ features and the corresponding reference points. The selected features and reference points are used as the initial queries and reference points of the detection head.
\subsection{The generality of our method}
To evaluate the generality of our method, we evaluate our method on three Transformer-base models including BEVFormer~\cite{li2022bevformer}, DETR3D~\cite{wang2022detr3d} and PETR~\cite{liu2022petr}. \\
\noindent\textbf{Implementation Details.} The implementation of \dasca is the same for all three methods, wherein we add depth-aware positional encodings to both image features and queries. However, the implementation of \dcl is slightly different because DETR3D and PETR do not have dense BEV feature maps like BEVFormer. We have described the implementation of \sdcl on these models in Section~\ref{subsec:DCL}.
\noindent\textbf{Results.} 
\begin{table*}
\centering
\setlength{\tabcolsep}{2.5pt}
\begin{tabular}{l|cccccc|c}
\toprule
\textbf{Method}                                    & \textbf{mAP}$\uparrow$  & \textbf{mATE}$\downarrow$ & \textbf{mASE}$\downarrow$  & \textbf{mAOE}$\downarrow$ & \textbf{mAVE}$\downarrow$ & \textbf{mAAE}$\downarrow$ & \textbf{NDS}$\uparrow$ \\
\midrule
BEVFormer  $^\dag$        & 0.416 & 0.673 & 0.274 & 0.372   & 0.394 & 0.198 & 0.517  \\
DAT-BEVFormer $^\dag$         & 0.428 & 0.633 & 0.273 & 0.331   & 0.327 & 0.188 &  \textbf{ 0.539\fontsize{7.0pt}{\baselineskip}\selectfont{(+2.2)}} \\
\midrule
DETR3D-R101-DCN           & 0.347 & 0.765 & 0.268 & 0.392   & 0.876 & 0.211 & 0.422  \\
DAT-DETR3D-R101-DCN            & 0.364 & 0.756 & 0.272 & 0.373   & 0.832 & 0.216 & \textbf{ 0.437\fontsize{7.0pt}{\baselineskip}\selectfont{(+1.5)}} \\
\midrule
PETR-V299$^*$             & 0.378 & 0.746 & 0.272 & 0.488   & 0.906 & 0.212 & 0.426  \\
DAT-PETR-V299$^*$        & 0.392 & 0.723 & 0.267 & 0.486   & 0.829 & 0.206 &\textbf{ 0.445\fontsize{7.0pt}{\baselineskip}\selectfont{(+1.9)}} \\
\bottomrule
\end{tabular}
\vspace{0.2cm}
\caption{The generality of our method. $^*$ denotes that the VoVNet-99 (V2-99) is pre-trained on depth estimation task with extra data~\cite{dd3d}. Note that models in this table are all single-stage models keep all settings the same except for our improvements and do not use CBGS.}
\label{tab:abl:univ}
\end{table*}
Table~\ref{tab:abl:univ} demonstrates that our DAT-BEVFormer significantly improves over BEVFormer by $+2.2$ NDS and $+1.2$ mAP.  Our method also improves DETR3D and PETR by $+1.5$ NDS and $+1.9$ NDS, respectively. These results verified the generality of our method and confirm our intuition that the lack of depth information is a common problem in DETR-based models like DETR3D and PETR.
\begin{table*}
\centering
\resizebox{0.8\textwidth}{!}{
\setlength{\tabcolsep}{1.5pt}
\begin{tabular}{l|c|r|cccccc|c} 
\toprule
\textbf{Method}             &  \textbf{Modality} &
\textbf{GFLOPs} &
\textbf{mAP}$\uparrow$  & \textbf{mATE}$\downarrow$ & \textbf{mASE}$\downarrow$  & \textbf{mAOE}$\downarrow$ & \textbf{mAVE}$\downarrow$ & \textbf{mAAE}$\downarrow$ & \textbf{NDS}$\uparrow$ \\
\midrule
CenterPoint-Voxel~\cite{yin2021center}       & L &-       & 0.564 &   -   &    -   &     -    &     -  &    -   & 0.648  \\
CenterPoint-Pillar        & L   &-     & 0.503 &   -    &    -   &     -    &  -     &     -  & 0.602  \\ 
\midrule
FCOS3D~\cite{wang2021fcos3d}      & C   &2008     & 0.295 & 0.806 & 0.268 & 0.511   & 1.315 & 0.170 & 0.372  \\
DETR3D$^\dag$~\cite{wang2022detr3d}      & C  & 1017      & 0.303 & 0.860 & 0.278 & 0.437   & 0.967 & 0.235 & 0.374  \\
BEVDet-R50~\cite{huang2021bevdet}          & C  & 215      & 0.286 & 0.724 & 0.278 & 0.590   & 0.873 & 0.247 & 0.372  \\
BEVDet-Base         & C  & 2963      & 0.349 & 0.637 & 0.269 & 0.490   & 0.914 & 0.268 & 0.417  \\
PETR-R50~\cite{liu2022petr}                & C  & -      & 0.313 & 0.768 & 0.278 & 0.564   & 0.923 & 0.225 & 0.381  \\
PETR-R101       & C   & -     & 0.357 & 0.710 & 0.270 & 0.490   & 0.885 & 0.224 & 0.421  \\
PETR-Tiny       & C   & -     & 0.361 & 0.732 & 0.273 & 0.497   & 0.808 & 0.185 & 0.431  \\
PETRv2~\cite{liu2022petrv2} & C   & -     & 0.401 & 0.745 & 0.268 & 0.448   & 0.394 & 0.184 & 0.496  \\
BEVDet4D-Tiny~\cite{huang2022bevdet4d}      & C   & 222     & 0.323 & 0.674 & 0.272 & 0.503   & 0.429 & 0.208 & 0.453  \\
BEVDet4D-Base$^*$      & C & 2989       & 0.421 & 0.579 & 0.258 & 0.329   & 0.301 & 0.191 & 0.545  \\
BEVDepth-R50      & C  & -      & 0.351 & 0.639 & 0.267 & 0.479   & 0.428 & 0.198 & 0.475  \\
BEVDepth$^*$     & C    & -    & 0.412 & 0.565 & 0.266 & 0.358   & 0.331 & 0.190 & 0.535  \\
BEVFormer  $^\dag$    & C    & 1304    & 0.416 & 0.673 & 0.274 & 0.372   & 0.394 & 0.198 & 0.517  \\
\midrule

DAT-BEVFormer $^{\S \dag}$    & C    & 1323    & \textbf{0.433} & 0.623 & 0.271 & 0.351   & 0.309 & 0.188 & \textbf{0.545}  \\
\bottomrule
\end{tabular}
}
\vspace{0.2cm}
\caption{Comparison on the nuScenes \emph{val} set. Without specification, the models use ResNet101~\cite{resnet} (R101) as the backbone. $^\dag$ denotes that the model adopts ResNet101-DCN~\cite{dai2017deformable} as backbone. L denotes LiDAR and C denotes camera. $^\S$ denotes the model uses rescale-based query selection (see~\ref{implementation}). $^*$ denotes the model uses CBGS~\cite{zhu2019class}.}
\label{tab:val}
\end{table*}
\begin{table*}
\vspace{-2mm}
\centering
\resizebox{0.8\textwidth}{!}{

\setlength{\tabcolsep}{2.5pt}
\begin{tabular}{l|c|c|cccccc|c}
\toprule
\textbf{Method}                                   & \textbf{Modality} &\textbf{Backbone} & \textbf{mAP}$\uparrow$  & \textbf{mATE}$\downarrow$ & \textbf{mASE}$\downarrow$  & \textbf{mAOE}$\downarrow$ & \textbf{mAVE}$\downarrow$ & \textbf{mAAE}$\downarrow$ & \textbf{NDS}$\uparrow$ \\
\midrule
CenterPoint~\cite{yin2021center}                              & L &-       & 0.674 & 0.255 & 0.235 & 0.339   & 0.233 & 0.128 & 0.718  \\ \midrule
FCOS3D~\cite{wang2021fcos3d}                                   & C  &R101      & 0.358 & 0.690 & 0.249 & 0.452   & 1.434 & 0.124 & 0.428  \\
DETR3D~\cite{wang2022detr3d}                                   & C &V2-99$^\dag$       & 0.412 & 0.641 & 0.255 & 0.394   & 0.845 & 0.133 & 0.479  \\
BEVDet-Pure~\cite{huang2021bevdet}                              & C &-        & 0.398 & 0.556 & 0.239 & 0.414   & 1.010 & 0.153 & 0.463  \\
BEVDet-Beta                              & C   &-      & 0.422 & 0.529 & 0.236 & 0.396   & 0.979 & 0.152 & 0.482  \\
PETR~\cite{liu2022petr} & C &V2-99$^\dag$        & 0.434 & 0.641 & 0.248 & 0.437   & 0.894 & 0.143 & 0.481  \\
PETR-e                                   & C   &Swin-T      & 0.441 & 0.593 & 0.249 & 0.384   & 0.808 & 0.132 & 0.504  \\
BEVDet4D$^*$~\cite{huang2022bevdet4d}                                 & C    &-     & 0.451 & 0.511 & 0.241 & 0.386   & 0.301 & 0.121 & 0.569  \\
BEVFormer~\cite{li2022bevformer}                         & C   &V2-99$^\dag$      & 0.481 & 0.582 & 0.256 & 0.375   & 0.378 & 0.126 & 0.569  \\ 
PETRv2~\cite{liu2022petrv2}
& C   &V2-99$^\dag$      & 0.490 & 0.561 & 0.243 & 0.361   & 0.343 & 0.120 & 0.582  \\ 
BEVDepth$^*$~\cite{li2022bevdepth}                                 & C   &V2-99$^\dag$      & 0.503 & 0.445 & 0.245 & 0.378   & 0.320 & 0.126 & 0.600  \\
\midrule
DAT-BEVFormer$^\S$                                & C   &V2-99$^\dag$      & \textbf{0.515} & 0.532 & 0.253 & 0.347   & 0.322 & 0.121 & \textbf{0.600}  \\
\bottomrule
\end{tabular} }
\vspace{0.2cm}
\caption{Comparison on the nuScenes \emph{test} set. L denotes LiDAR and C denotes camera. $^*$ denotes that the model uses CBGS. $\dag$ denotes that the VoVNet-99 (V2-99) is pre-trained on depth estimation task with extra data~\cite{dd3d}. $^\S$ denotes the model uses rescale-based query selection.}
\label{tab:test}
\end{table*}
\subsection{Comparison with top-performed methods}
For a fair comparison with our baseline model BEVFormer and other top-performing 3D detectors, we report our performance with R$101$-DCN backbone on nuScenes \texttt{val} in Table~\ref{tab:val}, which is the most popular setting reported by most works. We also provide a strong LiDAR-based method, CenterPoint~\cite{yin2021center}, for reference. According to the results, Additionally, compared to other models, our \modelname-BEVFormer has the best performance. BEVDet4D-Base has the same NDS as \modelname-BEVFormer, but its GFLOPS is much higher than ours, and its mAP is lower than ours.

To demonstrate the scalability of our model to stronger backbones, we report \modelname-BEVFormer with a pre-trained VoVNet-99 backbone on nuScenes \texttt{test} in Table~\ref{tab:test}. Without any additional techniques, our model outperforms the previous state-of-the-art (SOTA) model, BEVDepth, with an mAP that is +1.2 higher than theirs. Moreover, compared to BEVFormer, \modelname-BEVFormer shows an improvement of +3.4 mAP and +3.1 NDS.

\begin{table*}
\centering
\setlength{\tabcolsep}{2.5pt}
\begin{tabular}{c|l|cccccc|c}
\toprule
\textbf{Row}&
\textbf{Method}                                   & \textbf{mAP}$\uparrow$  & \textbf{mATE}$\downarrow$ & \textbf{mASE}$\downarrow$  & \textbf{mAOE}$\downarrow$ & \textbf{mAVE}$\downarrow$ & \textbf{mAAE}$\downarrow$ & \textbf{NDS}$\uparrow$ \\
\midrule
1.&BEVFormer              & 0.416 & 0.673 & 0.274 & 0.372   & 0.394 & 0.198 & 0.517  \\
2.&Row1+DA-SCA                                      & 0.419 & 0.660 & 0.272 & 0.345   & 0.353 & 0.189 & 0.527 \\
3.&Row2+\sdcl                                  & 0.428 & 0.633 & 0.273 & 0.331   & 0.327 & 0.188 & 0.539  \\

\bottomrule
\end{tabular}
\vspace{0.2cm}
\caption{Effectiveness of each component. The tale shows that both methods \dasca and \dcl effectively improve performance.}
\label{tab:abl:components}
\end{table*}
\begin{table*}
\centering
\setlength{\tabcolsep}{2.5pt}
\resizebox{0.7\textwidth}{!}{%
\begin{tabular}{c|l|cccccc|c}
\toprule
\textbf{Row}&
\textbf{Method}                                   & \textbf{mAP}$\uparrow$  & \textbf{mATE}$\downarrow$ & \textbf{mASE}$\downarrow$  & \textbf{mAOE}$\downarrow$ & \textbf{mAVE}$\downarrow$ & \textbf{mAAE}$\downarrow$ & \textbf{NDS}$\uparrow$ \\
\midrule
1.&Depth-wise Negative Suppression                               & 0.428 & 0.633 & 0.273 & 0.331   & 0.327 & 0.188 & 0.539  \\
2.&Random Negative Suppression  & 0.423 & 0.653 & 0.275 & 0.360   & 0.323 & 0.190 & 0.534  \\
3.&No Negative Suppression          & 0.420 & 0.663 & 0.269 & 0.377   & 0.359 & 0.185 & 0.525 \\
\midrule

4.&DNS in first $2$ layers                                       & 0.422 & 0.661 & 0.274 & 0.348  & 0.334 & 0.197 & 0.530  \\
5.&DNS in first $4$ layers                                      & 0.430 & 0.642 & 0.274 & 0.379   & 0.338 & 0.184 & 0.533  \\
6.&DNS in all 6 layers                               & 0.428 & 0.633 & 0.273 & 0.331   & 0.327 & 0.188 & 0.539  \\
\bottomrule
\end{tabular} }
\caption{Ablation on \dcl. Rows 1 to 3 evaluate the effectiveness of different approaches to sampling negative examples. Row 4 to 6 evaluate the effectiveness of applying \dcl to multiple layers of spatial cross-attention.}
\label{tab:abl:DCL}
\end{table*}
\subsection{Ablation study}

\begin{figure*}[h!]
    \centering
\includegraphics[width=0.8\linewidth]{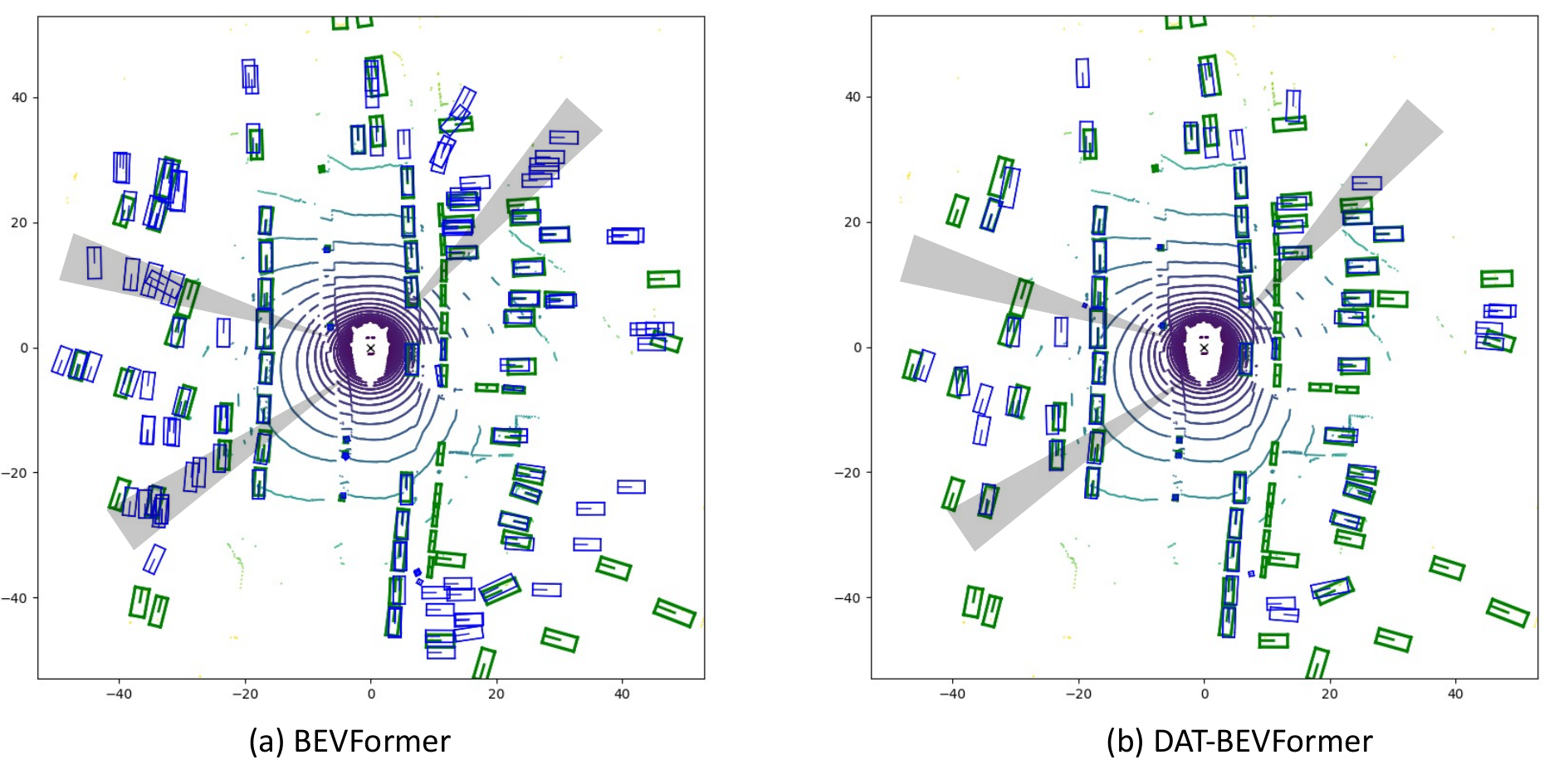}
    \caption{A visualization of predictions on BEV. Green boxes are GT boxes and blue ones are predicted ones. We use the same threshold of score $> 0.1$ to filter predictions for both BEVFormer and \modelname-BEVFormer. (a) The predictions enclosed by shaded triangles are duplicate predictions caused by the ambiguity problem shown in Section~\ref{sec:motivation}.(b) The prediction of the same sample by \modelname. The problem of duplicate predictions along depth axes is alleviated.
    }
    \vspace{0.2cm}
    \label{fig:vis}
\end{figure*}

Table~\ref{tab:abl:components} demonstrates the effectiveness of each component of our model individually. Row 1 presents our baseline model, BEVFormer. Rows 2 to 4 correspond to the addition of one component each. The inclusion of depth information into Spatial Cross-Attention improves NDS by 1.0, and conducting \dcl further improves NDS by 1.2.

\textbf{Impact on NDS.} Table~\ref{tab:abl:DCL} provides a more specific ablation on our key component, \dcl. The first three rows demonstrate the effectiveness of sampling negative examples along the object rays. For these three rows, we keep all other settings constant and only change the way negative examples are sampled. Note that we use DA-SCA for all results in Table~\ref{tab:abl:DCL}. It is evident that sampling depth-wise negative examples is better than sampling negative examples randomly.

\textbf{Impact on Multi-Layer NDS.} Rows 4 to 6 show that conducting \sdcl on multiple BEV encoder layers is beneficial. We have a total of 6 encoder layers, and it is evident that NDS increases the performance consistently as the number of layers conducting \sdcl increases.

\subsection{Visualization}
Figure~\ref{fig:vis} displays a visualization of the predictions made by our \modelname-BEVFormer, as well as the baseline model BEVFormer. To avoid duplication of predictions along the depth axes, Fig.\ref{fig:vis} (b) reduces these instances. This visualization serves to confirm that our model can indeed mitigate the ambiguity problem discussed in Section\ref{sec:motivation}.

However, there are still some issues with the predictions made by \modelname-BEVFormer. Specifically, when the object rays of two different objects overlap, the object in front typically has an accurate prediction, but the one behind may be missed or have a low-quality prediction. Two potential reasons may contribute to this problem. Firstly, the object behind may be obstructed by the one in front, making it difficult for both models to predict accurately. Secondly, in our \sdcl, two objects on the same depth axis may become negative examples of each other, leading to missed predictions.

\section{Conclusion}
\label{sec:conclusion}

In this paper, we identified a problem with previous Transformer-based 3D detectors that neglected depth information in the Spatial Cross-Attention, resulting in inaccurate and duplicated predictions along the depth axes. To address these issues, we proposed our Depth-Aware Spatial Cross-Attention (DA-SCA), which incorporates depth information into both BEV queries and image features. To further tackle the problem of duplicated predictions, we introduced the \dcl to explicitly reject predictions with incorrect depths.

Our experimental results demonstrate that our \modelname\ achieves significant performance improvements over all three baseline methods (BEVFormer, DETR3D, and PETR), confirming the generality of our approach. The ablation study confirms the effectiveness of each component of our model. Furthermore, our visualization results show that our model effectively resolves the lack-of-depth-awareness problem in previous Transformer-based 3D detectors.

\noindent\textbf{Limitations and Future Work:} Our current work focuses on improving Transformer-based 3D detectors using our proposed approach. However, depth learning remains a challenging problem for all camera-based 3D detectors, and we plan to investigate the application of our method to other detectors in the future. Additionally, we did not compare implicit depth learning with explicit depth learning in this paper. We aim to conduct further research on the differences between the two approaches and explore ways to combine them in the future.

{\small
\bibliographystyle{ieee_fullname}
\bibliography{sec/11_references}
}


\end{document}


\title{---Supplementary Materials---\\  Introducing Depth into Transformer-based 3D Object Detection}

\author{First Author\\
Institution1\\
Institution1 address\\
{\tt\small firstauthor@i1.org}
\and
Second Author\\
Institution2\\
First line of institution2 address\\
{\tt\small secondauthor@i2.org}
}

\maketitle
\ificcvfinal\thispagestyle{empty}\fi

\appendix
\setcounter{figure}{5}
\setcounter{figure}{5}
\appendix
\label{sec:appendix}
\begin{figure}[h!]
    \centering
\includegraphics[width=0.8\linewidth]{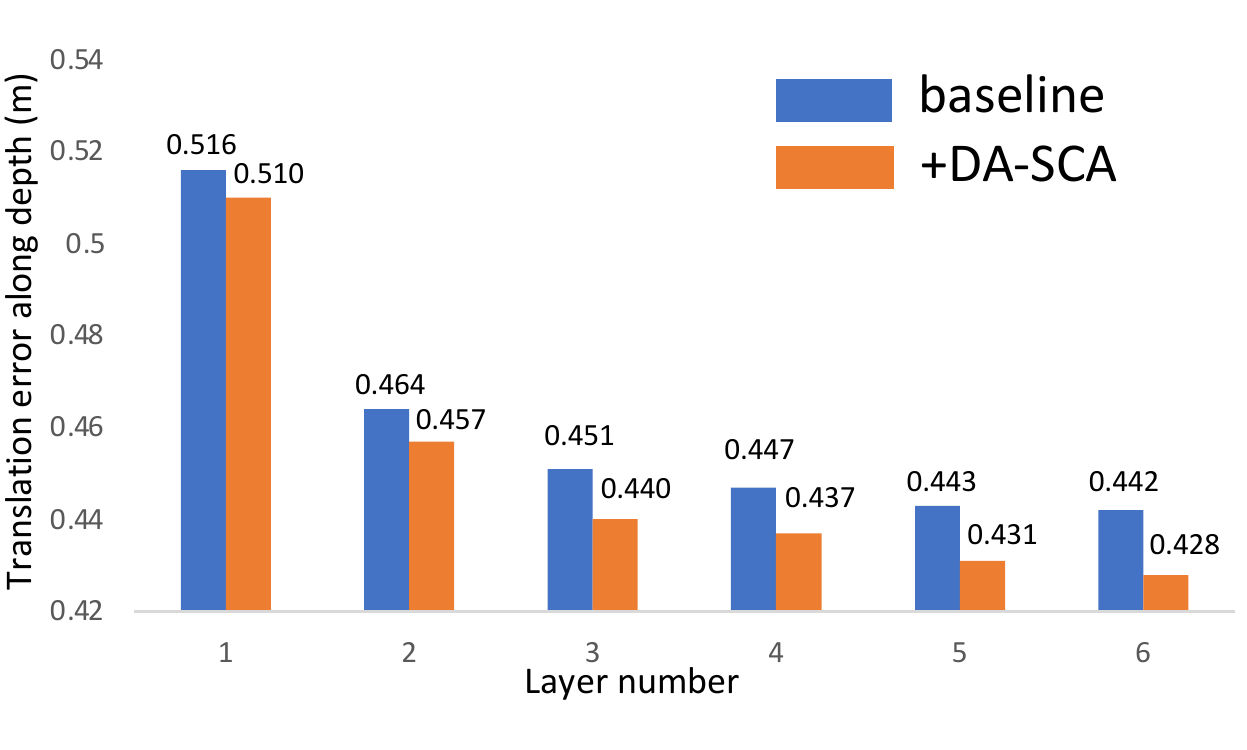}
    \caption{
        A comparison of BEVFormer and BEVFormer+DA-SCA in translation error along depth in different decoder layers. The x-axis is the layer number and the y-axis is the translation error along depth in meters.
    }
    \vspace{-0.2cm}
    \label{fig: trans dpt}
\end{figure}

\begin{figure*}
    \centering
\includegraphics[width=0.8\linewidth]{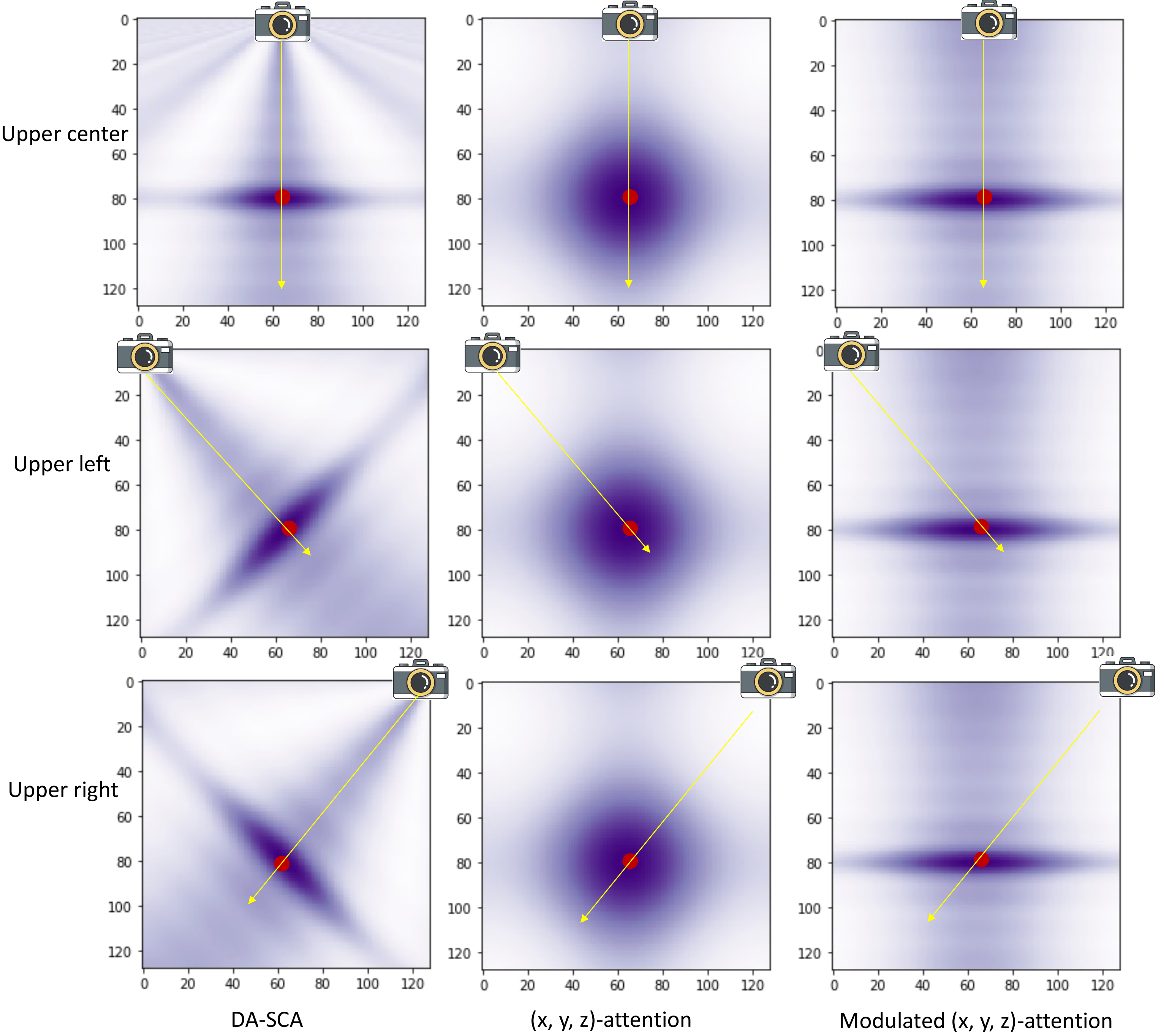}
    \caption{Illustration of the attention map of the $(u, v, d)$ positional encoding (PE) in DA-SCA and the $(x, y, z)$ PE in standard cross-attention from different camera views. The red point denotes the reference point and the yellow ray denotes the depth axis. The position of the camera is marked with an icon.
    }
    \label{fig: attn compare}
\end{figure*}

\section*{Overview}
This supplementary material presents more details and additional results not included in the main paper due to page limitation. The list of items included are:

\begin{itemize}
    \item More analyses of the proposed DA-SCA mechanism in Sec.~\ref{sec:supp_anl}.
    \item More experimental details in Sec.~\ref{sec:supp_imp}.
    \item More visual results of our dataset in Sec.~\ref{sec:supp_viz}.

\end{itemize}

\section{Module Analysis}
\label{sec:supp_anl}
In this section, we analyze how our \dasca works.

\noindent\textbf{Attention map.}
In Fig.\ref{fig: attn compare}, we visualize the attention maps of the $(u, v, d)$ positional encoding (PE) in DA-SCA, the $(x, y, z)$ PE in standard cross-attention, and the modulated $(x, y, z)$ PE as in\cite{liu2021dab}. To map the attention scores onto the BEV plane, we average the scores along the $z$ axis, and we indicate the reference point with coordinates $(x, y)$ in red. We show three different camera positions: upper center, upper left, and upper right. Compared to the $(x, y, z)$ PE, the attention map of our DA-SCA is more sensitive to changes along the depth axis in all three cases. While the modulated $(x, y, z)$ PE has a similar attention map to DA-SCA when the camera is in the upper center, its attention map does not change as the relative position of the camera with the reference point changes. Therefore, its attention map is not always sensitive along depth as our DA-SCA.
\\
\noindent\textbf{Translation errors along depth.} 
In Fig.\ref{fig: trans dpt}, we compare the translation errors along depth between BEVFormer\cite{li2022bevformer} and BEVFormer+DA-SCA across six decoder layers. The results show that BEVFormer+DA-SCA consistently outperforms BEVFormer in terms of translation errors along depth. Specifically, the mean translation error along depth for BEVFormer+DA-SCA is $0.014m$ smaller than that of BEVFormer. This difference is similar to the gap in the total mean translation error of $0.013m$ reported in Table 4, which supports the conclusion that DA-SCA mainly improves the model's performance along depth. Additionally, as the decoder layers become deeper, the performance gain of BEVFormer+DA-SCA becomes more significant, indicating that incorporating more DA-SCA layers leads to larger improvements.
\section{Implementation details}
\label{sec:supp_imp}
This section provides more implementation details of the proposed method and experiments.
\subsection{Depth-Aware Spatial Cross-Attention}
For Depth-Aware Spatial Cross-Attention (DA-SCA), we add sine positional encoding of $(u, v, d)$ in both image features and queries. Since we only detection the objects within the scope $x\in \left[-51.2, 51.2\right]$ and $y\in \left[-51.2, 51.2\right]$, the maximum $d$ is $51.2*\sqrt{2}=72.4$. Therefore we have $d\in \left[0, 72.4 \right]$. Because sine positional encoding requires $(u, v, d)$ normalized into $[0, 1]$, we use $(u/w,v/h,d/72.4)$ as the normalized coordinates for queries, where $w, h$ are the width and height of the image. For $d$ from the output of the depth network, we normalize it to $\left[0, 1\right]$ through sigmoid.
\subsection{Depth-aware Negative Suppression loss}
When conducting \dcl (\sdcl), we need to sample positive and negative examples along object rays. The object rays are rays that connect cameras and object centers. For simplicity, we use the BEV center as an approximation of the position of the cameras. For a GT object with center coordinates $(x, y)$ on BEV. Since the BEV center have coordinates $(0, 0)$, the direction of the object ray is approximately $(x-0, y-0)= (x, y)$ assuming the camera is at $(0, 0)$. In our experiments, we set the threshold for negative examples as $0.2$ and denote the coordinates of negative examples as $(\lambda_{neg}x,\lambda_{neg}y)$. Therefore, we have $\lambda_{neg} \in (0.0, 0.8) \cup (1.2, \frac{51.2}{max(x,y)})$.
\subsection{Our model with VoVNet-99 backbone}
We include the details of our model with VoVNet-99 backbone pre-trained in DD3D~\cite{dd3d}. We used multi-scale image augmentation during training. We randomly select a scale from $0.8$ to $1.2$ for each frame of images during training. Moreover, we also adopt Look Forward Twice proposed in ~\cite{zhang2022dino}, which allows gradients to pass to $(i-1)$-th decoder layer from $i$-th decoder layer. As the layer goes deeper, the gap becomes larger which verifies that more DA-SCA layers lead to better prediction in depth.
\begin{figure*}[tp]
    \centering
    \includegraphics[width=0.85\linewidth]{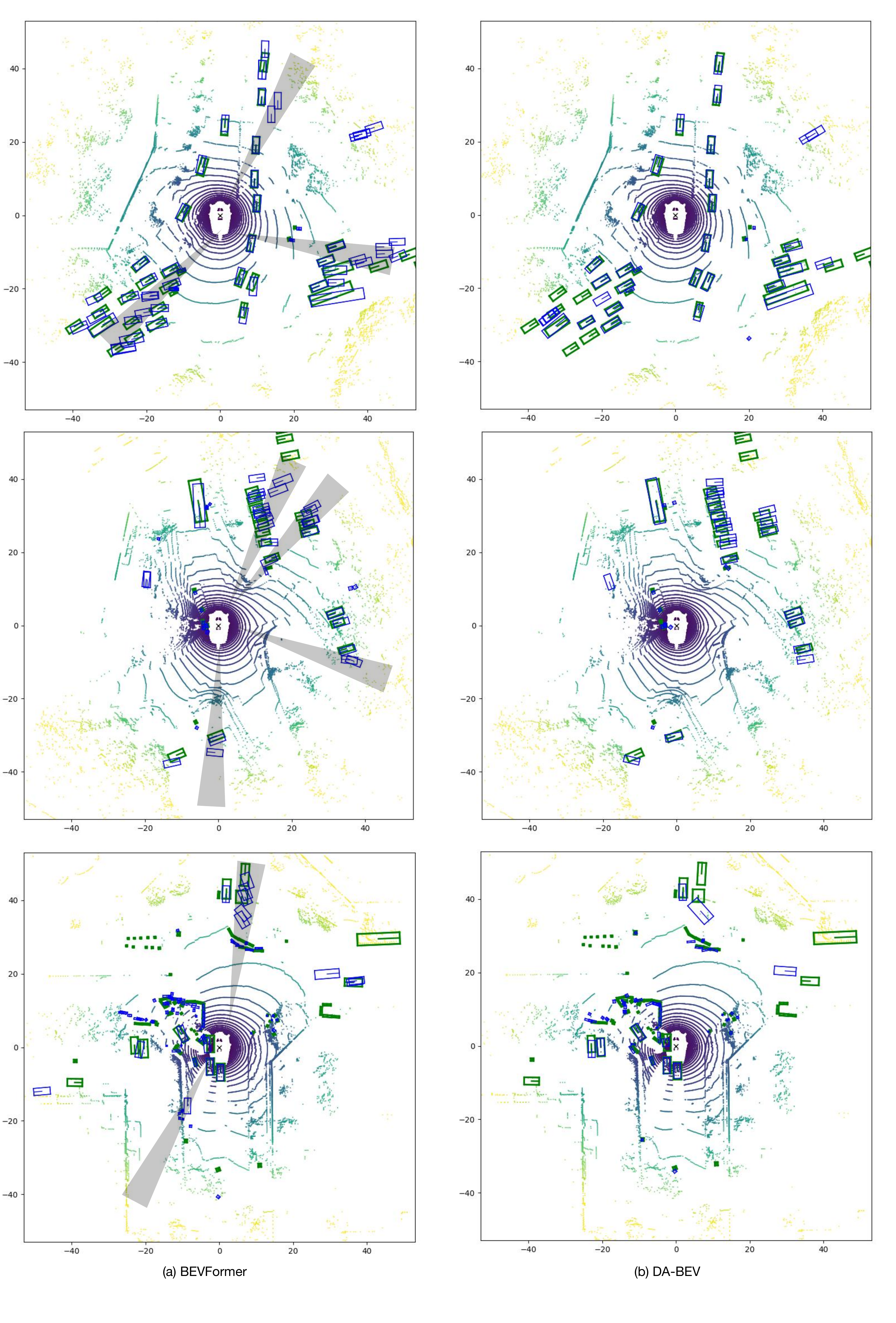}
    \vspace{-0.8cm}
    \caption{Visualization of the detection results on BEV of our model and BEVFormer. Green boxes are GT boxes, and blue ones are predicted ones. We use the same score threshold $> 0.1$ to filter predictions for both BEVFormer and DAT. (a) The predictions enclosed by shaded triangles are duplicate predictions caused by the ambiguity problem. (b) The prediction of the same sample by DAT. The problem of duplicate predictions along depth axes is alleviated.
    }
    \label{fig:appendix vis compare}
\end{figure*}

\section{Visualization}
\label{sec:supp_viz}
\subsection{Comparison with BEVFormer}
In Fig.~\ref{fig:appendix vis compare}, we give more examples of the predictions of our method compared with BEVFormer. These examples consistently verify that our method can alleviate the ambiguity problem.

\subsection{More visualizations}
\begin{figure*}[tp]
    \centering
    \includegraphics[width=1\linewidth]{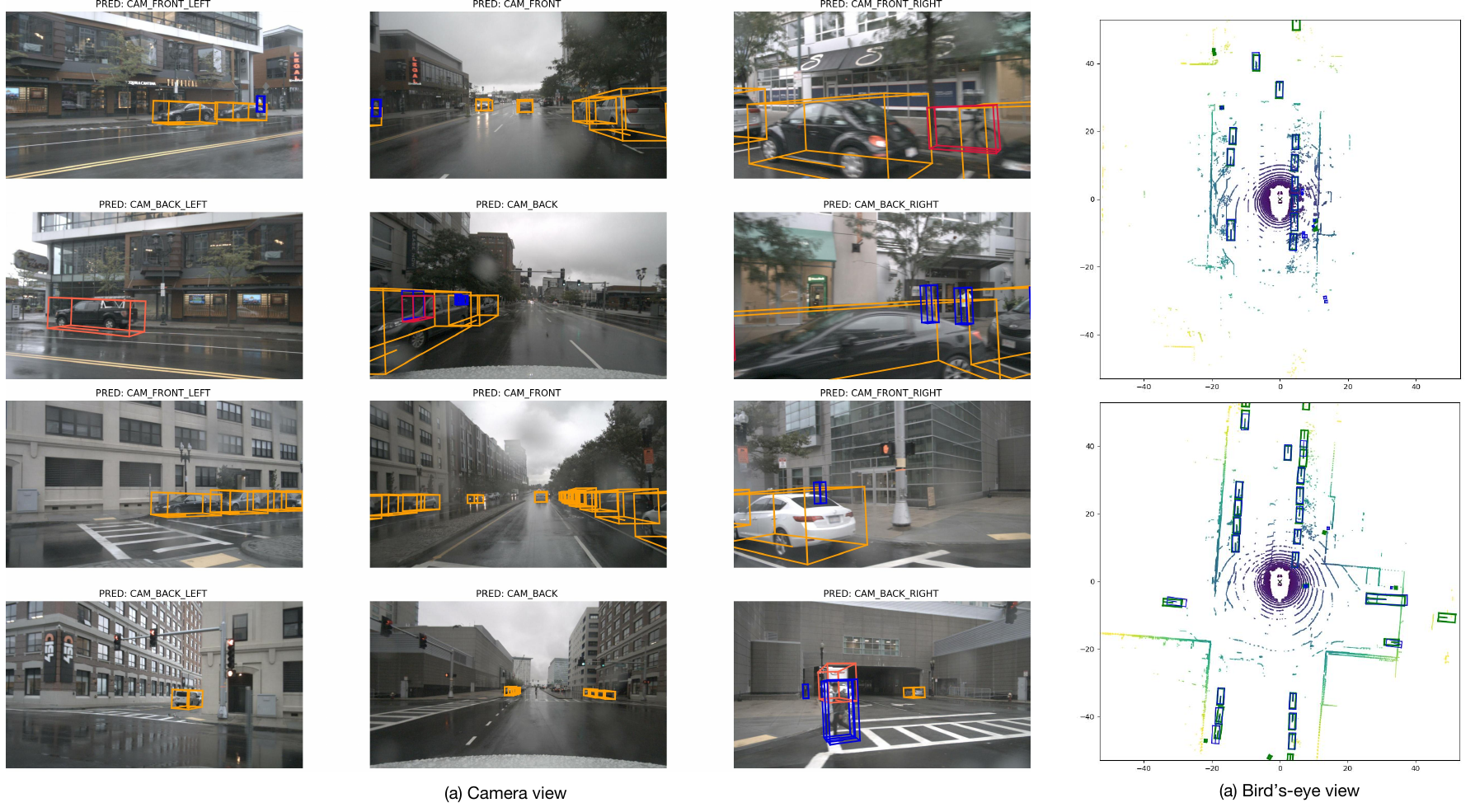}
    \caption{More detection results on both camera view and Bird's-eye view (BEV). (a) The first two rows are predictions of one frame on 6 cameras. The last two rows are predictions of another frame. (b) The green and blue boxes denote GT and predicted boxes, respectively.
    }
    \label{fig:appendix vis more}
\end{figure*}

In Fig.~\ref{fig:appendix vis more}, we show more detection results of \modelname.
{\small
\bibliographystyle{ieee_fullname}
\bibliography{sec/11_references}
}
